\pdfoutput=1

\documentclass[11pt]{article}

\usepackage{ACL2023}

\usepackage{times}
\usepackage{latexsym}

\usepackage[T1]{fontenc}

\usepackage[utf8]{inputenc}

\usepackage{microtype}

\usepackage{inconsolata}

\usepackage{microtype}
\usepackage{amsmath}
\usepackage{booktabs}
\usepackage{array}
\newcolumntype{L}{>{$}l<{$}}
\newcolumntype{C}{>{$}c<{$}}
\newcolumntype{R}{>{$}r<{$}}

\usepackage{enumitem}
\newcommand{\model}[1]{{\textsc{COFFEE}}}

\usepackage[utf8]{inputenc}

\usepackage{multirow}
\usepackage{array, tabularx, caption, boldline}
\usepackage{graphicx}
\usepackage{cellspace}
\usepackage{algpseudocode}  
\usepackage{amsmath}
\usepackage{multicol}  
\usepackage{multirow} 
\usepackage{flexisym}
\usepackage{graphicx}  
\usepackage{array}

\usepackage{microtype}
\usepackage{tcolorbox}

\usepackage{makecell}
\usepackage{latexsym}

\usepackage{multirow}
\usepackage{array, tabularx, caption, boldline}
\usepackage{graphicx}
\usepackage{cellspace}
\usepackage{algpseudocode}  
\usepackage{amsmath}
\usepackage{multicol}  
\usepackage{multirow} 
\usepackage{flexisym}
\usepackage{graphicx}  
\usepackage{array}

\usepackage{microtype}

\usepackage{tabularx}
\makeatletter
\def\hlinewd#1{%
\noalign{\ifnum0=`}\fi\hrule \@height #1 %
\futurelet\reserved@a\@xhline}
\makeatother

\usepackage{algpseudocode}  
\usepackage{amsmath}
\usepackage{multicol}  
\usepackage{multirow} 
\usepackage{graphicx}  
\usepackage{array}
\usepackage{arydshln}
\usepackage{booktabs}
\usepackage{textcomp}

\usepackage{amssymb}
\usepackage{pifont}
\usepackage{xcolor,colortbl}

\usepackage{times}
\usepackage{latexsym}
\usepackage{booktabs}
\usepackage{microtype}

\usepackage{soul}

\title{\model{}: A Contrastive Oracle-Free Framework for Event Extraction}

\author{Meiru Zhang \\
  \texttt{mz468@cam.ac.uk} \\\And
  Yixuan Su \\
  University of Cambridge \\
  \texttt{ys484@cam.ac.uk} \\\And 
  Zaiqiao Meng \\
  University of Glasgow \\
  \texttt{zaiqiao.meng.glasgow.ac.uk} \\\And
  Zihao Fu \\
  University of Cambridge \\
  \texttt{zf268@cam.ac.uk} \\\And 
  Nigel Collier \\
  University of Cambridge \\
  \texttt{nhc30@cam.ac.uk} \\}

\author{Meiru Zhang$^{\spadesuit}$\ \ \ \  \textbf{Yixuan Su}$^{\spadesuit}$\ \ \ \   \textbf{Zaiqiao Meng}$^{\spadesuit \diamondsuit}$\\ \textbf{Zihao Fu}$^\spadesuit$\ \ \ \  \textbf{Nigel Collier}$^\spadesuit$ \\
$^\spadesuit$Language Technology Lab, University of Cambridge \\
$^\diamondsuit$School of Computing Science, University of Glasgow \\
 \texttt{$^\spadesuit$\{mz468, ys484, zf268, nhc30\}@cam.ac.uk} \\
 \texttt{$^\diamondsuit$zaiqiao.meng@glasgow.ac.uk}
 }

\begin{document}
\maketitle
\begin{abstract}
Event extraction is a complex task that involves extracting events from unstructured text. Prior classification-based methods require comprehensive entity annotations for joint training, while newer generation-based methods rely on heuristic templates containing oracle information such as event type, which is often unavailable in real-world scenarios. In this study, we consider a more realistic task setting, namely the Oracle-Free Event Extraction (OFEE) task, where only the input context is given, without any oracle information including event type, event ontology, or trigger word. To address this task, we propose a new framework, \model{}. This framework extracts events solely based on the document context, without referring to any oracle information. In particular, \model{} introduces a contrastive selection model to refine the generated triggers and handle multi-event instances. Our proposed \model{} outperforms state-of-the-art approaches in the oracle-free setting of the event extraction task, as evaluated on two public variants of the ACE05 benchmark. The code used in our study has been made publicly available\footnote{\url{https://github.com/meiru-cam/COFFEE}}.
\end{abstract}

\section{Introduction}

\begin{figure}[t]
    \centering
    \includegraphics[width=\columnwidth]{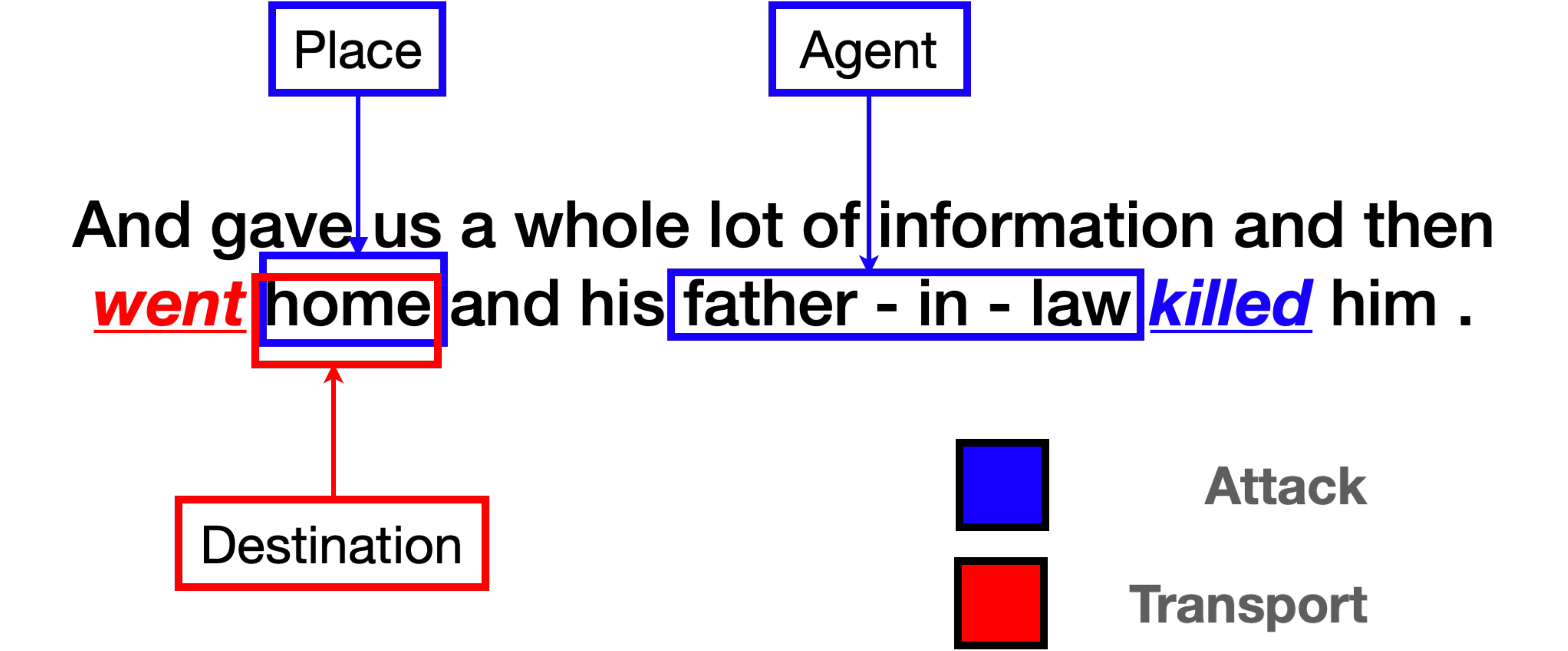}
    \caption{An event extraction example with two events: \textbf{Transport} and \textbf{Attack}. In the `Transport' event, `\underline{\textit{\textcolor{red}{went}}}' is the trigger word, and `home' is the `Destination' argument. In the `Attack' event, `\underline{\textit{\textcolor{blue}{killed}}}' is the trigger word while `father-in-law' and `home' are the `Agent' and `Place' arguments, respectively.}
    \vspace{-1em}
    \label{fig:example}
\end{figure}

The event extraction task aims to identify events and their arguments from the given textual input context~\cite{nguyen2016joint, wadden2019entity, yang2019exploring}. Conventionally, this task can be decomposed into four sub-tasks~\cite{nguyen2016joint}: (i) detecting the trigger word that most directly describes the event; (ii) event type classification for defining its event-specific attributes; (iii) argument identification and (iv) argument classification that maps the argument entities to the corresponding role attributes based on the structure of each event type, namely event schema. For instance, Figure \ref{fig:example} shows the input context of an event extraction example that contains two events: a `Transport' event triggered by the trigger word `went' and an `Attack' event triggered by the trigger word `killed', where `Transport' and `Attack' are two event types. 

 Many prior studies formulate the event extraction task as a token-level classification problem, which extracts event triggers and arguments using sequence tagging models based on tailor-designed neural networks~\citep{nguyen2016joint,liu2018gru, li2019biomedical,yang2019exploring, wadden2019entity, huang2020biomedical,lin2020joint,van2021cross}. However, such methods cannot leverage rich label semantics since the target outputs (e.g., event triggers and arguments) are fixed tagging labels. Recently, with advances in generative pre-trained language models, several generation-based approaches~\citep{hsu2022degree, huang2022multilingual, li2021document, zhang2021contrastive} have been applied to solve this problem. These approaches transform the event extraction task into a conditional generation task. By utilizing the autoregressive generation nature of generative pre-trained language models (e.g., BART-Gen~\citep{li2021document}, DEGREE~\citep{hsu2022degree}) and some manual prompts, it becomes possible to harness the semantics of labels and conduct both entity extraction and classification in an autoregressive manner simultaneously.

While impressive results are reported, we identify two major limitations of the current generation-based event extraction methods. Firstly, most of these methods rely on heuristic templates and extensive human knowledge engineering. According to the experiments conducted by~\citet{hsu2022degree}, a slight change in the template might lead to significant performance changes, thus raising the issue of using sub-optimal templates. Secondly, most of these generation-based approaches still require certain oracle information, such as event type and event schema, which necessitate extensive manual annotations. For example, the DEGREE model's inference process, as demonstrated by~\citet{hsu2022degree}, requires manually designed event-specific templates for each example and iterates over all event types. On the other hand, Text2Event~\citep{lu2021text2event} also constrains the generation with manually designed templates, which require event schema to be given. However, obtaining this oracle information, such as event type and schema, is unrealistic for a real-world inference system to achieve automatically. Hence, this paper aims to address the Oracle-Free Event Extraction (OFEE) task where only the input context is given.

In this study, we propose a novel Contrastive Oracle-Free Framework for Event Extraction (\model{}), which addresses the event extraction task without using any oracle information. Our \model{} consists of two parts, a generator that performs the extraction of events and a selector that aims to refine the generated results. The generator of our \model{} generates both the candidate triggers and event arguments, where the shared generator allows for cross-task knowledge sharing between these sub-tasks. The selector of our \model{} learns to re-rank and select the candidate triggers to obtain more accurate trigger predictions, which is inspired by~\citep{su2021few}. One challenge of the sentence-level event extraction is that a sentence may contain more than one event record~\citep{si2022generating, subburathinam2019cross} (e.g., the example in Figure \ref{fig:example}), and event specific templates can help the model to identify and extract events in a targeted manner. Prior approaches tackling this challenge have necessitated either multi-label tagging~\citep{ramponi2020biomedical, lin2020joint}, event-specific templates~\citep{hsu2022degree}, or multi-turn question answering techniques~\citep{du2020qaevent,li2020qaevent}. In contrast, our proposed model can concurrently generate and select multiple event candidates, encompassing both the event trigger and its associated type, thereby effectively addressing the aforementioned challenge. 

The contribution of this work is as follows:
\begin{itemize}[leftmargin=20pt, noitemsep, topsep=5pt, parsep=4pt, partopsep=3pt]
    \item We highlight the challenge of the current event extraction task setting and introduce the oracle-free setting of this task that requires the model to produce the structural event without using oracle information beyond the context.
    \item We propose \model{}, a novel {C}ontrastive {O}racle-{F}ree {F}ramework for {E}vent {E}xtraction which use a generator and a selector to generatively obtain structural event information from context without using any oracle information.
    \item We conduct experiments on two variants of the ACE05 benchmark under the oracle-free setting to evaluate our \model{}. The results demonstrate that the template-based baselines heavily rely on the additional oracle information, whereas our \model{} exhibits superior empirical performance over these baselines in the absence of an oracle.
\end{itemize}

\begin{figure*}
\centering
  \includegraphics[width=1.0\linewidth]{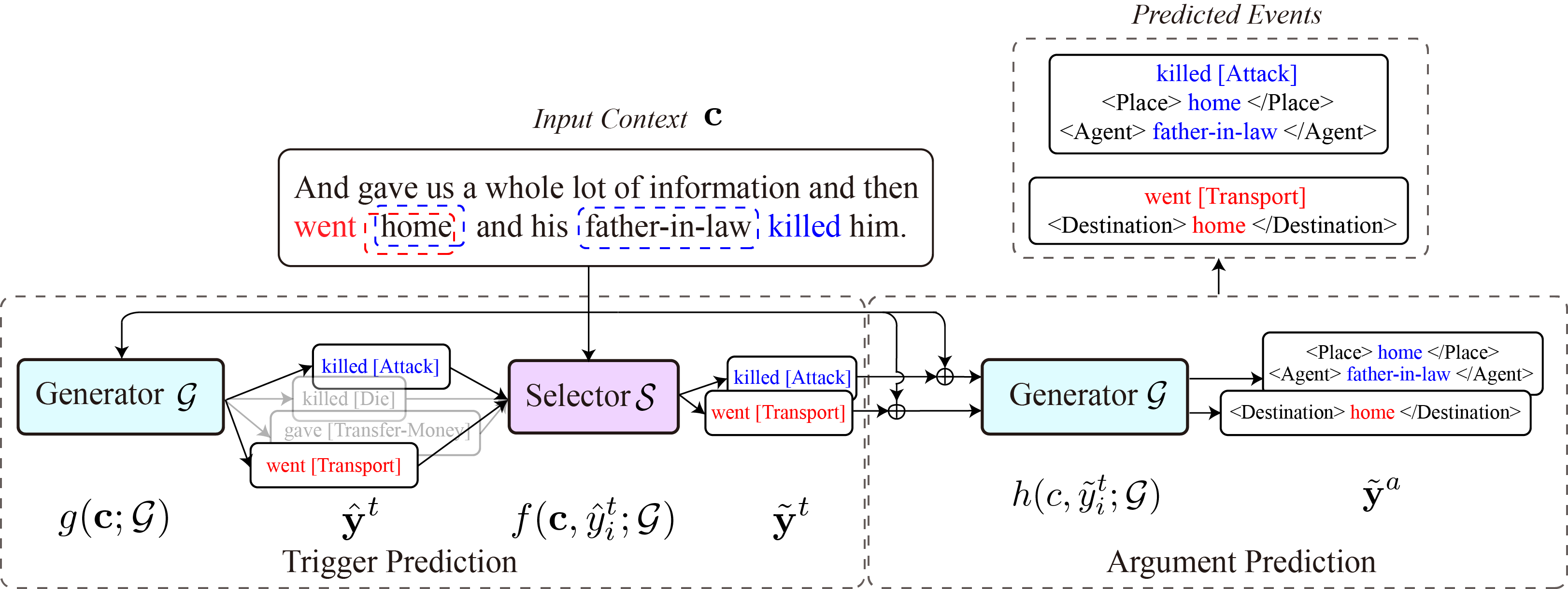}
  \caption{Overview of our proposed \model{} framework. We train $\mathcal{G}$ to generate trigger candidates $\hat{\mathbf{y}}^t$ that contain trigger word and event type first. These trigger candidates then used to train $\mathcal{S}$ to select the final trigger predictions $\tilde{\mathbf{y}}^t$. In the argument prediction stage, the trained generator is re-used to generate arguments $\tilde{\mathbf{y}}^a$ based on $\tilde{\mathbf{y}}^t$ selected by $\mathcal{S}$. Only the input context $\mathbf{c}$ is required to predict events.}
  \vspace{-1em}
  \label{fig:framework}
\end{figure*}

\section{Task Definition}
\label{sec:task}
Conventionally, the event extraction task entails the following terminologies~\cite{nguyen2016joint, liu2020qaevent, paolini2021structure}.
\begin{itemize}[leftmargin=10pt, noitemsep, topsep=5pt, parsep=4pt, partopsep=3pt]
    \item \emph{Input Context}: The input sentence or sentences that contain one or more events.
    \item \emph{Trigger Word}: The main word that most clearly expresses the occurrence of an event (e.g., words `went' and `killed' in Figure \ref{fig:example}).
    \item \emph{Event Type}: The event type that defines the semantic structure of a specific event (e.g., events `Transport' and `Attack' in Figure \ref{fig:example}).
    \item \emph{Event Argument}: Event arguments identify the entities involved in events and their roles based on their relationships with the event triggers. An entity can be an object, place or person that participates in the event. For example, `home' is an entity that serves as both the `Place' argument of the `Transport' event and the `Destination' argument of the `Attack' event in Figure \ref{fig:example}.
\end{itemize}

Given the input context $\mathbf{c}$, which is a sequence of tokens $[c_1, \cdots, c_n]$, the conventional event extraction task aims to identify the trigger words, classify the events triggered by these words and extract the arguments in each of the events with their corresponding roles~\citep{nguyen2016joint, chen2015event}. Assume that an input sentence context $\mathbf{c}$ contains $|e|$ different events, then its ground truth triggers $\mathbf{y}^{t}$ can be represented as $[y^{t}_1, \cdots, y^{t}_{|e|}]$, where $y^{t}_i$ denotes the $i$-th trigger word and event type of the given context sentence. For each event, there is a list of ground truth arguments, denoted by $\mathbf{y}^{a}$, which is a list of $\langle$\emph{role}, \emph{argument}$\rangle$ pairs, i.e., $\mathbf{y}^{a}=[\langle r_1, a_1 \rangle, \cdots, \langle r_m, a_m \rangle]$, where $a_j$ is the $j$-th entity participating in the event and $r_j$ is the corresponding role type for that entity.

For this conventional event extraction task, the current state-of-the-art generation-based approaches rely on manual templates, which require trigger words or event types to be given, to simplify this task~\citep{hsu2022degree, lu2021text2event}. However, in a realistic scenario, although argument roles are event-specific, gold trigger words or event type information may not be readily available during event argument extraction
We focus on the Oracle-Free Event Extraction (OFEE) task, which presents a more practical scenario by only providing the input context during inference. The goal of OFEE is to infer event triggers and arguments without relying on pre-defined event specific templates, making it more challenging to solve due to the absence of external guidance or oracle information.

\section{Methodology}\label{sec:model}
As mentioned in the task definition, our goal is to extract event frames without using any templates. This adds complexity to the generation model, particularly when dealing with contexts containing multiple events, such as the example given in Figure \ref{fig:example}. To address the challenging OFEE task, we propose a novel contrastive framework called \model{}, which comprises two primary components: a generator $\mathcal{G}$, responsible for generating event frames present in the provided context, and a selector $\mathcal{S}$, which re-ranks and selects the triggers generated by $\mathcal{G}$. In our proposed \model{} framework, $\mathcal{G}$ is fine-tuned using ground truth triggers and arguments (i.e. $\mathbf{y}^t$ and $\mathbf{y}^a$) to generate candidate triggers $\hat{\mathbf{y}}^t$ and arguments $\tilde{\mathbf{y}}^a$ (see \S \ref{subsec:generator}). At the inference stage, $\mathcal{S}$ is fine-tuned to refine and select final trigger predictions $\tilde{\mathbf{y}}^t$ based on the generated candidate triggers $\hat{\mathbf{y}}^t$ and gold triggers $\mathbf{y}$ on the training set (see \S \ref{subsec:selector}). The final trigger predictions are forwarded to $\mathcal{G}$ for argument prediction (see \S \ref{subsec:generator}). We next present the details of \model{}'s components, i.e. the generator and the selector.


\subsection{Generator}
\label{subsec:generator}
The generator is fine-tuned on both trigger prediction and argument prediction simultaneously by training on the pairs of instances with different prefixes `TriggerEvent: ' and `Argument: ' (see \S\ref{subsec:implementation}). In order to take the context as input and generate structured event frames, the generator $\mathcal{G}$ of \model{} is employed using an encoder-decoder transformer model, such as BART, T5 and mT5~\citep{lewis2020bart, raffel2020exploring, xue2021mt5}. We resort T5~\citep{raffel2020exploring} as the base model and encode only `[and]' and `[none]' as additional special tokens based on experimental results.

During the inference stage, we apply beam search~\citep{jelinek1976continuous} to generate candidate triggers $\hat{\mathbf{y}}^t$ and output the beam score of these triggers. Given the context $\mathbf{c}$, the generator outputs the top-$l$ triggers with the highest beam scores, denoted by $\hat{\textbf{y}}^t=g(\mathbf{c}; \mathcal{G})$, where $\hat{\textbf{y}}^t$ is a list of triggers $[\hat{y}^t_{1}, \cdots, \hat{y}^t_{l}]$, with beam scores $[b_1, \cdots, b_l]$, and $\hat{y}^t_{i}$ represents a generated candidate trigger in the context $\mathbf{c}$. After obtaining the list of candidate triggers, we use a contrastive-learning based selector $\mathcal{S}$ (see \S\ref{subsec:selector} for details) to further re-rank the generated candidates using $f(\mathbf{c},\hat{\textbf{y}}^t; \mathcal{S})$ and select the final set of trigger predictions $\tilde{\mathbf{y}}^t$. The predicted trigger words are then concatenated to the context iteratively, and the generator performs argument prediction on each event using $h(\mathbf{c}, \tilde{y}^t_i;\mathcal{G})$. Specifically, given the input $[c_1, \cdots, c_n, \tilde{y}^t_i]$, $G$ generates $\tilde{\mathbf{y}}^a=[\langle\tilde{r}_1, \tilde{a}_1\rangle, \cdots, \langle\tilde{r}_m, \tilde{a}_m\rangle]$ for a predicted trigger $\tilde{y}^t_i$.

\subsection{Selector}
\label{subsec:selector}
In our approach, we employ contrastive learning to re-rank the candidate triggers $\hat{\textbf{y}}^t$. Contrastive learning~\citep{chen2020simple} is a technique that aims to learn meaningful representations by maximizing the similarity between positive pairs while minimizing the similarity between negative pairs. In the context of our problem, we define the ground truth triggers $\textbf{y}^t$ for context $\mathbf{c}$ as the positive anchors, while the negative samples are the other incorrect candidates generated, i.e., $\hat{y}^t_{j} \not\in \mathbf{y}^t$. 

To apply contrastive learning for re-ranking, we first encode the context and candidate triggers using a shared encoder. 
Specifically, given a list of candidate triggers $\hat{\textbf{y}}^t$, for each $\hat{y}^t_{i}\in\hat{\textbf{y}}^t$ we concatenate it to the context and use $\mathcal{S}$ to map the concatenated text $[\mathbf{c}:\hat{y}^t_{i}]$ into a real-valued ranking score by preforming linear projection $f(\mathbf{c},\hat{y}^t_{i};\mathcal{S})$. In this study, we employ RoBERTa~\cite{liu2019roberta} as the backbone selector model to encode the text input, and $\mathcal{S}$ predicts the ranking score for each of the candidate triggers in $\hat{\mathbf{y}}^t$ through optimizing over a contrastive objective $\mathcal{L_S}$, which encourages $\mathcal{S}$ to predict higher scores for true trigger candidates and lower scores for false trigger candidates. 

Formally, given a context $\mathbf{c}$ and the generated candidate triggers $\hat{\mathbf{y}}^t$, $\mathcal{S}$ is fine-tuned to optimize:
\begin{equation}
    \small
    \mathcal{L}_{\mathcal{S}} = \sum_{i=1}^{|e|} \sum_{j=1}^k \max\{0, \rho -f(\mathbf{c}, y^t_i; \mathcal{S}) + f(\mathbf{c}, \hat{y}^t_{j}; \mathcal{S}) \},
\end{equation}
where $\hat{y}^t_{j} \not\in \mathbf{y}^t$, $\rho \in [-1, 1]$ is a pre-defined margin and $k$ represents the number of negatives sampled from $\hat{\mathbf{y}}^t$. By taking into account the implicit correlation between the context and generated candidates, $\mathcal{S}$ captures the semantic relevance between context and correct trigger candidates, thus enhancing trigger extraction and positively impacting the performance of argument extraction.

Since the number of events in the context is unknown, we use a threshold to automatically control the number of events predicted. Let $\alpha$ represent the weight parameter and $\theta$ represent the threshold parameter in our model. These hyperparameters are used for combining the beam score $b_i$ with the ranking score $s_i$ and filtering out the false candidate triggers, respectively. We determine the threshold $\theta$ and the weight $\alpha$ on the development set, which is exclusively utilized for hyperparameter tuning, to ensure an unbiased evaluation on the test set. The final set of trigger predictions is defined as $\tilde{\mathbf{y}}^t=\{\tilde{{y}}^t_{1}, \cdots, \tilde{{y}}^t_{|\tilde{\mathbf{y}}^t|}\}$, which satisfies that $\forall \tilde{{y}}^t_{i}$,
\begin{equation}
\small
\alpha\cdot\sigma(f(\mathbf{c}, \tilde{y}^t_{i}; \mathcal{S}))+(1-\alpha)\cdot\sigma (b_i) > \theta, 
\end{equation}
where $\sigma$ denotes the softmax function.

\section{Experiments}
\subsection{Dataset}\label{sec:dataset}
In this work, we evaluate our \model{} based on a public event extraction benchmark ACE05~\cite{walker2005ace}, which consists of 599 English documents, 33 event types, and 22 argument roles. Building upon previous works~\citep{wadden2019entity, lin2020joint} that split and preprocess  this dataset, we use two variants for the event extraction dataset, namely \textbf{ACE05-E} and \textbf{ACE05-E+}. Detailed split and statistics of the two datasets can be found in Table \ref{ACE05dataset}. 

\begin{table}
\centering
\resizebox{.47\textwidth}{!}{
\begin{tabular}{lllllll}
\toprule
\small
&&\multicolumn{2}{c}{\textbf{ACE05-E}}&\multicolumn{2}{c}{\textbf{ACE05-E+}}\\
\cmidrule(lr){3-4}
\cmidrule(lr){5-6}
 & \# sent & \# triggers & \# args & \# triggers & \# args \\
\hline

train & 17172 & 4202 & 4859 & 4419 & 6607 \\
val & 923 & 450 & 605 & 468 & 759 \\
test & 832 & 403 & 576 & 424 & 689 \\
\bottomrule
\end{tabular}
}
\caption{\label{ACE05dataset}The statistics of our used datasets.}
\vspace{-1em}
\end{table}



\subsection{Evaluation Metrics}
The evaluation of trigger identification, event type classification, argument identification, and argument role classification tasks utilizes the F1-score metric, consistent with the previous studies~\cite{zhang2019extracting, wadden2019entity}. A correct trigger classification prediction requires accurate trigger word and event type prediction, i.e., $\tilde{y}^t_i=y^t_i$. Correct argument identification necessitates accurate classification of the event type and argument entity, while a correct argument role classification demands accurate identification of the argument and role type prediction. Specifically, a predicted event type $\tilde{t}_e$, argument $\tilde{a}$, and role type $\tilde{r}$ are considered correct if $(\tilde{a}, \tilde{r}, \tilde{t}_e) = (a, r, t_e)$.

\subsection{Baselines}\label{sec:baseline}
To validate the effectiveness of our proposed method, we compared our \model{} with five state-of-the-art baselines: 
\begin{itemize}[leftmargin=10pt, noitemsep, topsep=5pt, parsep=4pt, partopsep=3pt]
    \item  \textbf{OneIE}~\cite{lin2020joint} is a joint neural model that simultaneously extracts entities and relations using a dynamic relation graph.
    \item \textbf{Text2Event}~\cite{lu2021text2event} is a sequence-to-structure controlled generation model with constrained decoding for event extraction. It focuses on the structured generation that uses event schema to form event records.
    \item \textbf{BART-Gen}~\cite{li2021document} is designed for document-level event extraction that can deal with the long-distance dependence issue and co-reference problem. Constrained generation is applied for argument extraction that requires event-specific templates.
    \item \textbf{DEGREE}~\cite{hsu2022degree} is a generative event extraction approach that highly relies on the designed template.
    \item \textbf{TANL}~\cite{paolini2021structure} is a model that extracts event triggers and arguments by so called augmented translation that embeds target outputs into the context sentence.
\end{itemize}

\subsection{Implementation}
\label{subsec:implementation}
We preprocess the data by separating original samples into event samples and inserting placeholders for target entities. The instances are processed with distinct prefixes for subtasks: `TriggerEvent: ' and `Arguments: '. Figure \ref{fig:datapreprocessing} shows a data preprocessing example. Details pertaining to our pipeline training and inference process, including specifics about the two-stage fine-tuning, such as the learning rate and batch size, as well as the beam search strategy employed during inference, are elaborated in Appendix~\ref{app:implementation-details}.

\begin{figure}[ht]
\centering
\begin{tcolorbox}[colback=green!5,colframe=blue!15,boxsep=0pt,left=0pt,right=0pt,top=1pt,bottom=1pt]
\scriptsize
\textbf{Input:}\\
"\textit{TriggerEvent:} And gave ... then went home ... killed him .",\\
"\textit{Arguments:} And gave ... then went home ... killed him . \textit{<Trigger>} killed",\\
"\textit{TriggerEvent:} And gave ... then went home ... killed him .",\\
"\textit{Arguments:} And gave ... then went home ... killed him. \textit{< Trigger>} went",
\end{tcolorbox}

\begin{tcolorbox}[colback=green!5,colframe=blue!15,boxsep=0pt,left=0pt,right=0pt,top=1pt,bottom=1pt]
\scriptsize
\textbf{Target:}\\
"killed [Life_Die]",\\
"\textit{<Agent>} father - in - law \textit{</Agent>} $\ldots$ \textit{<Place>} home \textit{</ Place >}",\\
"went [Movement_Transport]",\\
"\textit{<Artifact>} [ None] \textit{</Artifact>} $\ldots$ \textit{<Place>} [None] \textit{</Place>}",
\end{tcolorbox}
\caption{Example of input and target for the model.}
\vspace{-1em}
\label{fig:datapreprocessing}
\end{figure}


\section{Results}
\begin{table*}[t]
    \small
	\centering  
	\renewcommand{\arraystretch}{1.2}
	\setlength{\tabcolsep}{6pt}
	 \resizebox{.85\textwidth}{!}{
	\begin{tabular}{clcccccccc}
		\hlinewd{0.75pt}
	    &\multirow{2}{*}{\textbf{Model}}&\multicolumn{4}{c}{ACE-05E}&\multicolumn{4}{c}{ACE-05E+}\\
	    \cmidrule(lr){3-6}
	    \cmidrule(lr){7-10}
	    &&Trig I&Trig C&Arg I&Arg C&Trig I&Trig C&Arg I&Arg C\\\hline
            &OneIE&76.83&73.05&57.26&54.31&77.31&74.01&56.66&54.29\\\hline
	    &Text2Event$^\natural$ &73.93&69.06&51.59&49.52&73.40&68.99&52.64&50.39\\
	    &BARTGen$^\natural$ &74.36&71.13&55.22&53.71&-&-&66.62&64.28\\
	    &DEGREE$^{\natural\diamondsuit\flat}$ &74.57&70.96&56.03&53.41&74.90&70.30&55.74&53.61\\
	    \hlinewd{0.75pt}
	    \multirow{6}{*}{\rotatebox[origin=c]{90}{Oracle Free}}&Text2Event &73.49&68.60&51.24&49.32&72.73&68.30&52.48&50.35\\
	    &BARTGen &70.96&66.59&48.47&46.36&-&-&51.43&47.49\\
	    &DEGREE &43.64&2.18$^\dagger$&28.54&25.99&54.32&2.26$^\dagger$&30.09&28.79\\
	    &TANL&\textbf{81.10}&\textbf{77.09}&55.28&52.16&\textbf{80.28}&\textbf{76.03}&54.56&52.57\\
	    \cline{2-10}
	    &\textbf{COFFEE}&\underline{79.61}&\underline{75.73}&\textbf{59.88}&\textbf{55.43}&\underline{78.28}&\underline{74.70}&\underline{56.87}&\underline{54.11}\\
	    &\makecell[c]{\textit{+TANL}}&\textbf{81.10}&\textbf{77.09}&\underline{58.74}&\underline{55.24}&\textbf{80.28}&\textbf{76.03}&\textbf{59.78}&\textbf{57.06}\\
		\hlinewd{0.75pt}
	\end{tabular}
	}
    \caption{Performance comparison of \model{} and SOTA generation-based approaches. $^\dagger$ The trigger classification F1 of DEGREE is nearly zero because the model cannot exclude the negative samples constructed without a template.$^\natural$, $^\diamondsuit$, and $^\flat$ denote the model that requires a manually designed template, example keywords, and event description, respectively. The highest results are in \textbf{bold} and the second highest results are \underline{underlined}.}
    	\vspace{-1.5mm}
\label{tb:mainresult}
\end{table*}

\subsection{OFEE performance}
As described in Section \ref{sec:baseline}, Text2Event, BART-Gen and DEGREE utilize different oracle information. To compare the performance of our \model{} framework with these methods under the OFEE setting, we implemented the following adaptations to these baseline approaches:
\begin{itemize}[leftmargin=10pt, noitemsep, topsep=5pt, parsep=4pt, partopsep=3pt]
    \item Text2Event~\citep{lu2021text2event} relies on a complex constrained decoding mechanism that depends on the event schema. For the oracle-free setting, we utilized the default decoding of the T5 model to generate results.
    \item BART-Gen~\citep{li2021document} adopts a constrained generation mechanism, which necessitates the use of templates. We removed the template and the constrained decoding, thereby enabling the model to function. The trigger extraction performance of BART-Gen is not reported in our study due to an implementation error stemming from different preprocessing methods, which prevented us from applying this approach to the ACE-05E+ dataset. Consequently, we depended on the ground truth triggers for argument extraction in this instance.
    \item The DEGREE~\citep{hsu2022degree} model is designed to generate `invalid' instances during both the training and inference phases, wherein event-specific knowledge is combined with context even if no such event is mentioned in the context. We eliminated these event-specific templates, leaving only the context sentence as input.
\end{itemize}

As presented in Table \ref{tb:mainresult}, we report F1 scores of the compared methods over four sub-tasks described in \ref{sec:dataset}, namely trigger identification, trigger classification, argument identification, and argument classification. We observe the following:

\begin{itemize}[leftmargin=10pt, noitemsep, topsep=5pt, parsep=4pt, partopsep=3pt]
    \item Firstly, it is crucial to highlight that the oracle-free setting poses a more challenging scenario. When all oracle information is removed, generation-based baselines relying on templates exhibit a varying degree of performance decline on both datasets ($\downarrow$ 0.5\% to 37.42\% in argument classification). 
    Although DEGREE is effective with the oracle information, it struggles to filter out the `invalid' events in the oracle-free setting, resulting in an almost zero (2.18\%) trigger classification F1. This indicates that the information leaked in the template significantly contributes to the performance of DEGREE.
    \item Our proposed \model{} outperforms the classification-based approach OneIE and the generation-based approaches Text2Event, BARTGen, and DEGREE in both the presence and absence of oracle information across \textbf{all four metrics}. This demonstrates that our \model{} can effectively leverage the input context to extract event frames.
    \item In comparison to TANL, our \model{} achieves similar results in trigger extraction, with a difference of only 1.36\%. One possible explanation is that the threshold-based method results in a smaller recall value due to more false positives. However, our model possesses robust argument extraction capabilities and attained superior performance in argument extraction with these extracted triggers ($\uparrow$ 3.33\% and $\uparrow$ 2.46\% on ACE-05E and ACE05E+, respectively). These findings corroborate the effectiveness of the shared generator on trigger and argument prediction.

\end{itemize}

\subsection{Ablation study}
We conducted an ablation study on the threshold and weight parameters to demonstrate the effectiveness of our selector $\mathcal{S}$ and the influence of these parameters on the \model{} performance.

\begin{figure}[h]
\centering
\includegraphics[width=\columnwidth]{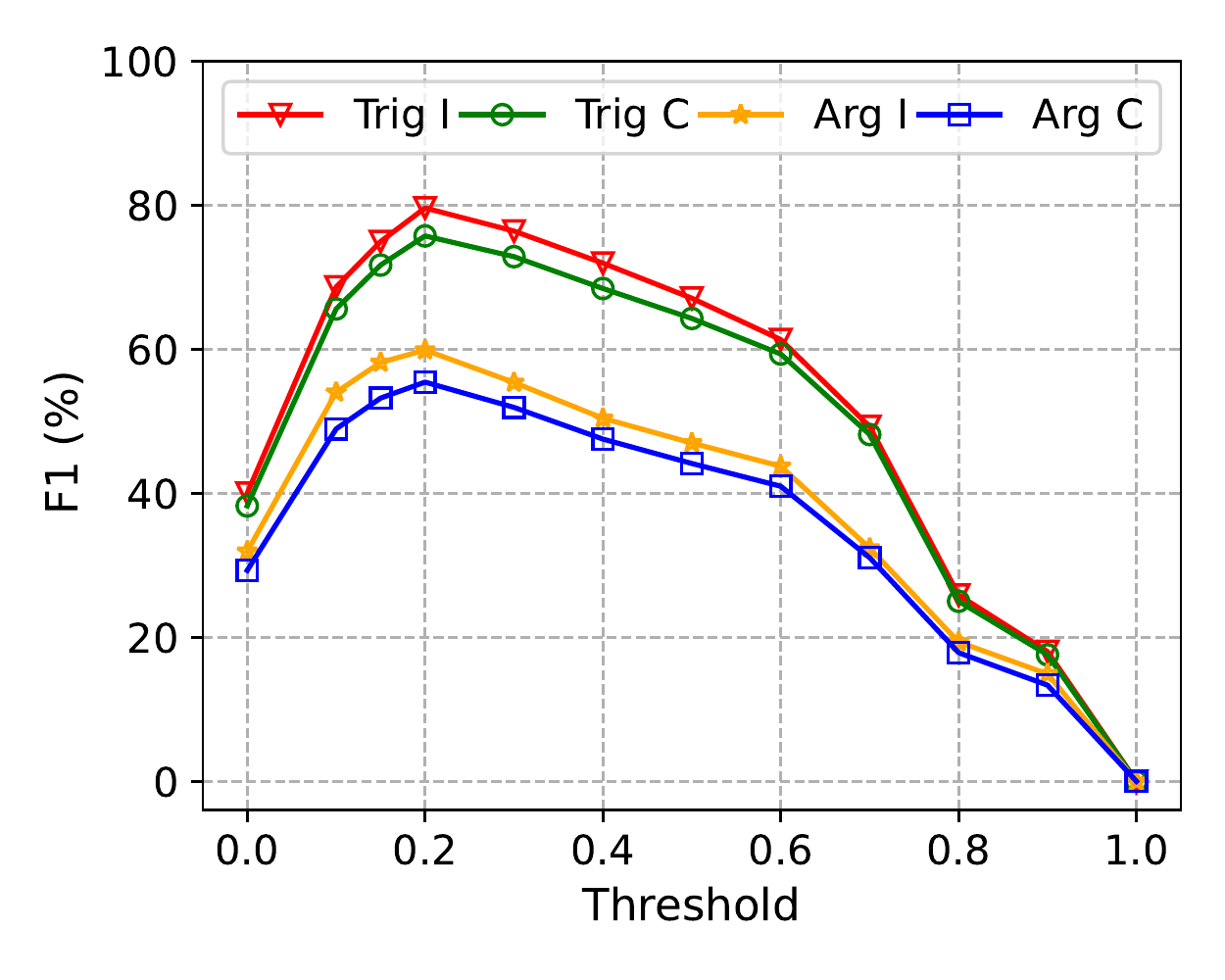}
\vspace{-2em}
\caption{Effect of threshold in \model{} framework.}
\label{fig:ablation_threshold}
\vspace{-1em} 
\end{figure}

Figure \ref{fig:ablation_threshold} illustrates the effect of the threshold parameter on \model{}. The threshold determines the minimum score a candidate must achieve to be selected. Increasing the threshold results in fewer candidates being selected but with higher accuracy. Conversely, an overly high threshold could filter out some of the correct candidates, decreasing performance. The optimal threshold value is 0.2, which achieves the best performance on all four subtasks.

\begin{figure}[h]
\centering
\includegraphics[width=\columnwidth]{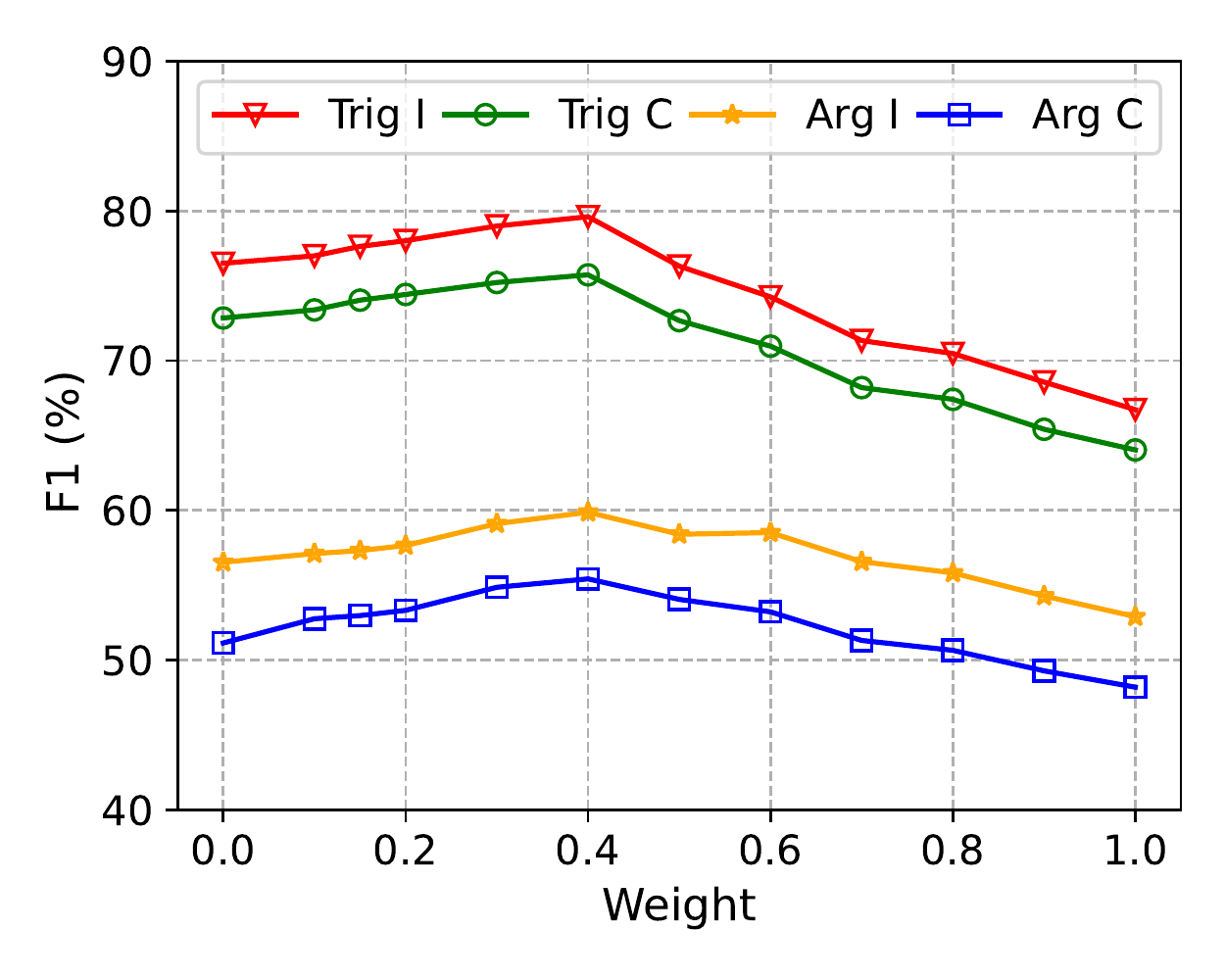}
\vspace{-2em}
\caption{The influence of the weight $\alpha$ on performance.}
\label{fig:ablation_weight}
\vspace{-1em} 
\end{figure}

In addition, Figure \ref{fig:ablation_weight} demonstrates the influence of the weight parameter on \model{}. The weight represents the ratio of combining the ranking score and generation score. When the weight is set to 0, only the generation score is considered, while a weight of 1 means that only the ranking score is considered. As depicted in Figure \ref{fig:ablation_weight}, the best extraction performance is achieved with a fixed threshold and an optimal weight value of $\alpha = 0.4$. The initial improvement in the F1 score with increasing weight suggests that the ranking score can effectively refine the results of the beam search. However, the ranking scores exhibit significant variations, leading to a corresponding fluctuation in softmax probability as the weight increases. As the final probability becomes increasingly reliant on the ranker probability, fewer candidates are selected at the same threshold, resulting in a decline in performance.


\subsection{Qualitative Case Analysis}
\begin{table*}[t]
    \centering  
    \renewcommand{\arraystretch}{1.2}
\resizebox{.85\textwidth}{!}{
\begin{tabular}{p{0.23\textwidth}|p{0.4\textwidth}|p{0.55\textwidth}}
\toprule
  \multicolumn{3}{c}{\textbf{Example 1}}\\\hline
  \textbf{Context} & \multicolumn{2}{p{\textwidth}}{Kommersant business daily joined in , declaring in a furious front - page headline : " The United States is demanding that Russia , France and Germany pay for the Iraqi war .} \\\hline
  \multirow[t]{2}{*}{\textbf{Reference}} 
  & \textbf{E1}: pay [Transfer-Money]  & \textbf{Args}: [Giver] Germany \\
  & \textbf{E2}: war [Attack] & \textbf{Args}: [Place] Iraqi \\\hline
  \multirow[t]{2}{*}{\textbf{TANL + COFFEE}} 
  & \textbf{E1}: & \textbf{Args}: \\
  & \textbf{E2}: war [Attack] & \textbf{Args}: [Place] Iraq \\\hdashline
  \multirow[t]{2}{*}{\textbf{\model{} w/o Ranker}} 
  & \textbf{E1}:  & \textbf{Args}: \\
  & \textbf{E2}: war [Attack] & \textbf{Args}: [Place] Iraq \\\hdashline
  \multirow[t]{2}{*}{\textbf{\model{}}} 
  &\colorbox{pink}{\textbf{E1}: pay [Transfer-Money]} & \colorbox{pink}{\textbf{Args}: [Giver] Germany} \\
  & \textbf{E2}: war [Attack] & \textbf{Args}: [Place] Iraq \\

  \midrule\midrule
  \multicolumn{3}{c}{\textbf{Example 2}}\\\hline
  \textbf{Context} & \multicolumn{2}{p{\textwidth}}{Welch specifically is seeking performance evaluations , correspondence between his estranged wife and partners while she worked at the law firm 's office in London , and documents related to her prospects of becoming a partner .} \\\hline
  \multirow[t]{2}{*}{\textbf{Reference}} 
  & \textbf{E1}: correspondence [Phone-Write] & \textbf{Args}: [Entity] partners; [Place] office \\
  & \textbf{E2}: becoming [Start-Position] & \textbf{Args}: [Entity] firm \\\hline
  \multirow[t]{2}{*}{\textbf{TANL + COFFEE}}
  & \textbf{E1}: correspondence [Phone-Write] & \textbf{Args}: [Entity] partners; [Place] office \\
  & \textbf{E2}: & \textbf{Args}: \\\hdashline  
  \multirow[t]{2}{*}{\textbf{\model{} w/o Ranker}} 
  & \textbf{E1}: & \textbf{Args}: \\
  & \textbf{E2}: becoming [Start-Position] & \textbf{Args}: \\\hdashline
  \multirow[t]{2}{*}{\textbf{\model{}}} 
  & \textbf{E1}: correspondence [Phone-Write] & \textbf{Args}: [Entity] partners; [Place] office \\
  & \colorbox{pink}{\textbf{E2}: becoming [Start-Position]} & \textbf{Args}: \\

\bottomrule
\end{tabular}
}
\caption{Event extraction examples from the test set using \model{}, \model{} without ranking and TANL+\model{}. The triggers and arguments missed by the baselines but captured by \model{} are \colorbox{pink}{highlighted}. It is evident that \model{} is generally more effective in detecting the events.}

\vspace{-0.5em}
\label{tb:case_study}
\end{table*}

In order to demonstrate the ability of our model to select event candidates, we analyze the results of two instances selected from the test set. For comparison, we select \model{} without ranking and TANL, given its high performance. As shown in Table \ref{tb:case_study}, our proposed model successfully extracts the missing events not detected by the baselines. The re-ranking mechanism enables the model to select more accurate candidates.

In particular, only \model{} successfully predicts all the events within the context. In Example 1, both TANL and \model{} without ranking fail to extract \textbf{E1}, triggered by `pay', suggesting that the baselines may have difficulty identifying complex event triggers. In this case, there is not a specific amount of money to be paid, but a mention of cost. In Example 2, TANL fails to extract \textbf{E2}, which is triggered by `becoming', and \model{} without ranking fails to extract \textbf{E1}, highlighting the inability of the baselines to identify events and their corresponding arguments consistently. In contrast, our \model{} successfully identifies the events and extracts the target arguments, demonstrating its superior performance.

Comparing \model{} with and without ranking, we can conclude that re-ranking in the selector is crucial. In both examples, \model{} fails to detect all events without re-ranking. Even though both candidates are the correct targets, the beam scores differ more than expected, which leads to incorrect ranking. The re-ranking can increase the probability of the second candidate and thus allowing it to be selected under the chosen threshold. 

These examples demonstrate the improvements in event extraction offered our selector $\mathcal{S}$, which allows the framework to re-rank and select the correct triggers for multi-event instances, outperforming the baselines and establishing our model as a more effective and reliable solution for OFEE tasks.


\section{Related Work}

\subsection{Event Extraction}
%

Early event extraction research primarily relied on rule-based methods involving hand-written patterns to identify event triggers and arguments in text~\citep{li2013joint, kai2015improving}. Supervised machine learning techniques became popular, with various feature-based classification models employed~\citep{hsi2016leveraging}. However, these methods faced limitations due to manual feature engineering and the need for large annotated datasets. Researchers then turned to deep learning approaches, utilizing convolutional neural networks (CNNs)~\citep{chen2015event, nguyen2015event, bjorne2018biomedical, yang2019exploring}, recurrent neural networks (RNNs)~\citep{nguyen2016joint}, and Tree-LSTM~\citep{li2019biomedical} for event extraction, which automatically learned relevant features and improved performance.

The introduction of pre-trained language models revolutionized event extraction. Fine-tuning these models achieved state-of-the-art performance across various benchmarks~\citep{lin2020joint, ramponi2020biomedical, wadden2019entity, yang2021document}. These models captured deep contextual information and benefited from knowledge transfer, enhancing performance with limited annotated data. Some studies framed event extraction as a multi-turn question answering task~\citep{du2020qaevent, li2020qaevent, liu2020qaevent, zhou2021role}, while others approached it as a sequence-to-sequence generation task~\citep{hsu2022degree, lu2021text2event, li2021document}. Although effective, these methods heavily relied on manually designed prompts and templates, except for Text2Event~\citep{paolini2021structure}, which depended solely on context information. In contrast, our work focuses on oracle-free event extraction and addresses the task via generation without templates. The most recent studies have focused on event detection or event argument extraction separately ~\citep{zhang2022zero, huang2022multilingual, ma2022prompt}, which is not directly comparable to our study as we consider the complete event extraction process. 

\subsection{Post-Generation Ranking}
Post-generation re-ranking is usually applied in two-stage systems, that is, generation and re-ranking, to re-score the output from the first stage by training an additional re-ranking module. This technique has been widely used in neural translation and summarization. For example,~\citet{ng2019facebook, yee2019simple} re-score and select the best hypotheses using Noisy Channel Modeling to improve translation quality.~\citet{zhong2020extractive} formulate the summarization as text matching and re-ranks the summary candidates based on similarity score.~\citet{liu2021simcls} introduce an additional scoring model with contrastive training to predict the score of generated summaries. Both methods utilize a margin-based ranking loss that initializes candidates with orders. For the event trigger selection, we assume that the beam score is not a reliable indicator and consequently treat the candidates equally.~\citet{su2021few} use a contrastive re-ranking module with hinge loss to select prototypes for a table-to-text generation. To the best of our knowledge, our work is the first to focus on enhancing the oracle-free generation-based event extraction models using re-ranking.

\section{Conclusion}
In this work, we study a more realistic setting of the event extraction task, namely the oracle-free event extraction, where no additional information beyond the context is required for event inference. To address this task, we propose a generation-based event extraction framework called \model{}. Our \model{} introduces a contrastive selector to improve trigger extraction performance by re-ranking and automatically determining the number of triggers to be selected in a given context. Additionally, we investigate the dependence of current generation-based models on extra knowledge, such as designed event-specific templates, event trigger keywords, and event descriptions. Our results show that this reliance on templates and human-designed trigger sets is unnecessary, and a pure oracle-free model applied directly can perform very well on general event extraction. In the future, we plan to extend sentence-level event extraction to document-level and explore zero-shot settings to handle the emergence of unseen events.

\clearpage

\section{Limitations}
Despite its promising results, our study has limitations. Our model primarily works with English text, limiting its applicability to other languages. Its focus on sentence-level extraction doesn't consider document context, which could be investigated in future research. The employed training dataset is relatively small, potentially not encompassing all possible event types, thus affecting the model's performance and generalizability. Additionally, our two-stage inference framework, while enhanced by a ranking module, is prone to error propagation. If a trigger isn't identified in the first stage, its associated arguments cannot be extracted. Future work should address these issues for improved performance and broader applicability.

\section{Ethics Statement}
In preparing and submitting this research paper, we affirm that our work adheres to the highest ethical standards and is devoid of any ethical issues. The study presented in the manuscript was conducted in a manner that respects the principles of academic integrity, transparency, and fairness.

\bibliography{acl2023}
\bibliographystyle{acl_natbib}

\clearpage
\appendix

\section{Appendices}
\label{sec:appendix}

\subsection{Implementation Details}
\label{app:implementation-details}
Our pipeline training comprises two stages: generation model fine-tuning and re-ranking model fine-tuning. The T5-base model~\cite{raffel2020exploring} fine-tuning is achieved through the HuggingFace Transformers library~\cite{wolf2019huggingface} on an RTX3090 GPU, using an AdamW optimizer~\cite{adamw}, with a learning rate of 0.0001 with a decay schedule of 1e-5. We set a batch size of 8 and maximum input/output sequence lengths at 650/200.

For inference, we generate candidate triggers using a beam search strategy with 10 beams. These candidates are then re-ranked and filtered by the selector $\mathcal{S}$, based on optimal thresholds and weights derived through grid search.

In the second stage, we fine-tune a RoBERTa-base model~\cite{liu2019roberta} for re-ranking. This stage reduces the maximum input length to 512 and sets the number of negative candidates for contrastive learning to 5, with a learning rate of 0.005.

Upon refining triggers, they are concatenated to the context and reintroduced to the generator for argument generation via a greedy search. The final extraction of entities and roles is conducted using regular expressions. Our \model{} generator and selector take approximately 4 hours and 2 hours to train, respectively.

Performance evaluation of the trigger and argument extraction is based on regular expressions to detect entities extracted from the placeholders. The results can be observed in Table \ref{tb:mainresult}.

\end{document}